\documentclass[lettersize,journal]{IEEEtran}
\usepackage{amsmath,amsfonts}
\usepackage{algorithmic}
\usepackage{algorithm}
\usepackage{array}
\usepackage[caption=false,font=normalsize,labelfont=sf,textfont=sf]{subfig}
\usepackage{textcomp}
\usepackage{stfloats}
\usepackage{url}
\usepackage{verbatim}
\usepackage{graphicx}
\usepackage{cite}
\usepackage{booktabs}
\usepackage{hyperref}
\usepackage[T1]{fontenc}
\usepackage{makecell}
\usepackage{xspace}
\usepackage{subfig}

\begin{document}

\title{CBEN -- A Multimodal Machine Learning Dataset for Cloud Robust Remote Sensing Image Understanding}

\author{Marco Stricker, Masakazu Iwamura, Koichi Kise}



\maketitle

\begin{abstract}
	Clouds are a common phenomenon that distorts optical satellite imagery, which poses a challenge for remote sensing. However, in the literature cloudless analysis is often performed where cloudy images are excluded from machine learning datasets and methods. Such an approach cannot be applied to time sensitive applications, e.g., during natural disasters. A possible solution is to apply cloud removal as a preprocessing step to ensure that cloudfree solutions are not failing under such conditions. But cloud removal methods are still actively researched and suffer from drawbacks, such as generated visual artifacts. Therefore, it is desirable to develop cloud robust methods that are less affected by cloudy weather. Cloud robust methods can be achieved by combining optical data with radar, a modality unaffected by clouds. While many datasets for machine learning combine optical and radar data, most researchers exclude cloudy images. We identify this exclusion from machine learning training and evaluation as a limitation that reduces applicability to cloudy scenarios. To investigate this, we assembled a dataset, named CloudyBigEarthNet (CBEN), of paired optical and radar images with cloud occlusion for training and evaluation. Using average precision (AP) as the evaluation metric, we show that state-of-the-art methods trained on combined clear-sky optical and radar imagery suffer performance drops of 23-33 percentage points when evaluated on cloudy images. We then adapt these methods to cloudy optical data during training, achieving relative improvement of 17.2-28.7 percentage points on cloudy test cases compared with the original approaches. Code and dataset are publicly available at: \url{https://github.com/mstricker13/CBEN}
\end{abstract}


\section{Introduction}
\label{sec:intro}



Clouds cover around 55\% of Earth’s land surface at any given time and are therefore, obfuscating satellite captured imagery based on optical sensors\cite{king2013spatial}. Their captured information about the ground is reduced to either a complete loss of information or corrupted information at pixels where clouds or cloud shadows are appearing\cite{zhai2018cloud}. These artifacts can confuse automated algorithms. Cloud shadows are darkening areas which then resemble water or burned areas more closely\cite{marshak2006impact}, while clouds themselves mimic snowy regions\cite{ma2023cloud}. Furthermore, clouds partially occlude spatial features of objects, essentially changing how they are represented as an input. As a result, the appearance of clouds negatively impacts automated optical based remote sensing analysis\cite{gawlikowski2022explaining}. 

Common approaches to overcome the appearance of clouds are cloudless analysis, cloud removal and cloud robust analysis. 

Cloudless analysis, the standard approach utilized in the literature, systematically avoids cloudy imagery by selecting images that have been captured on a day close to a reference date where the cloud cover is below a small threshold \cite{wang2023ssl4eo,stewart2023ssl4eo,bonafilia2020sen1floods11,helber2019eurosat,sumbul2021bigearthnet,velazquez2025earthview}. Cloudy images are avoided for two reasons. First, they are more difficult to label and if only optical images are available, then it is impossible to extract information from below the clouds. Secondly, the overall performance in multi-modal models decreases when optical images are corrupted by clouds, as this is the most relevant modality for machine learning solutions\cite{wang2023ssl4eo,shen2023evaluating,he2025optical,crowson2019comparison}. However, this is problematic because around 55\% of land surface are covered by clouds \cite{king2013spatial}, raising the question whether machine learning methods that perform well on clear-sky imagery retain their performance under realistic, cloudy conditions \cite{wilson2016remotely,schmitt2023there}. Maintaining a high degree of performance under varying weather conditions is critical for applications where it is not possible to wait for a cloud free observation. This includes tasks in regions with persistent cloud cover\cite{prudente2022multisensor}, real time event monitoring for illegal activity detection, such as logging, mining or fishing\cite{thompson2021preventing,suresh2013change,harinivashini2024optimizing,ding2024monitoring}, or disaster management during e.g., floods\cite{stricker2023fusing,miyamoto2023effect}, earthquakes\cite{zhao2021advances} or wildfires\cite{mohapatra2022early}

Therefore, applying cloud removal as a preprocessing step for cloudless analysis is a possible solution, that is however still an ongoing research suffering from the following drawbacks\cite{ebel2023uncrtaints}. Firstly, pairing cloudy input data  with cloud free target data is only possible by utilizing images at different points in time and therefore introducing pixel level inconsistencies between the target image and the real ground truth\cite{zhang2024lightweight}. Additionally, thin and thick clouds require different processing strategies and call for accurate detection of cloud edges and associated shadows \cite{zhang2024lightweight}. Then, bright surfaces such as snow can be confused with clouds \cite{ma2023cloud}. Lastly, reconstructed regions may exhibit visible artifacts or altered brightness relative to their surroundings \cite{ma2023cloud}. 

\begin{figure} 
	\centering
	\subfloat[\label{subfig:optical_cloudfree}]{%
		\includegraphics[width=0.45\linewidth]{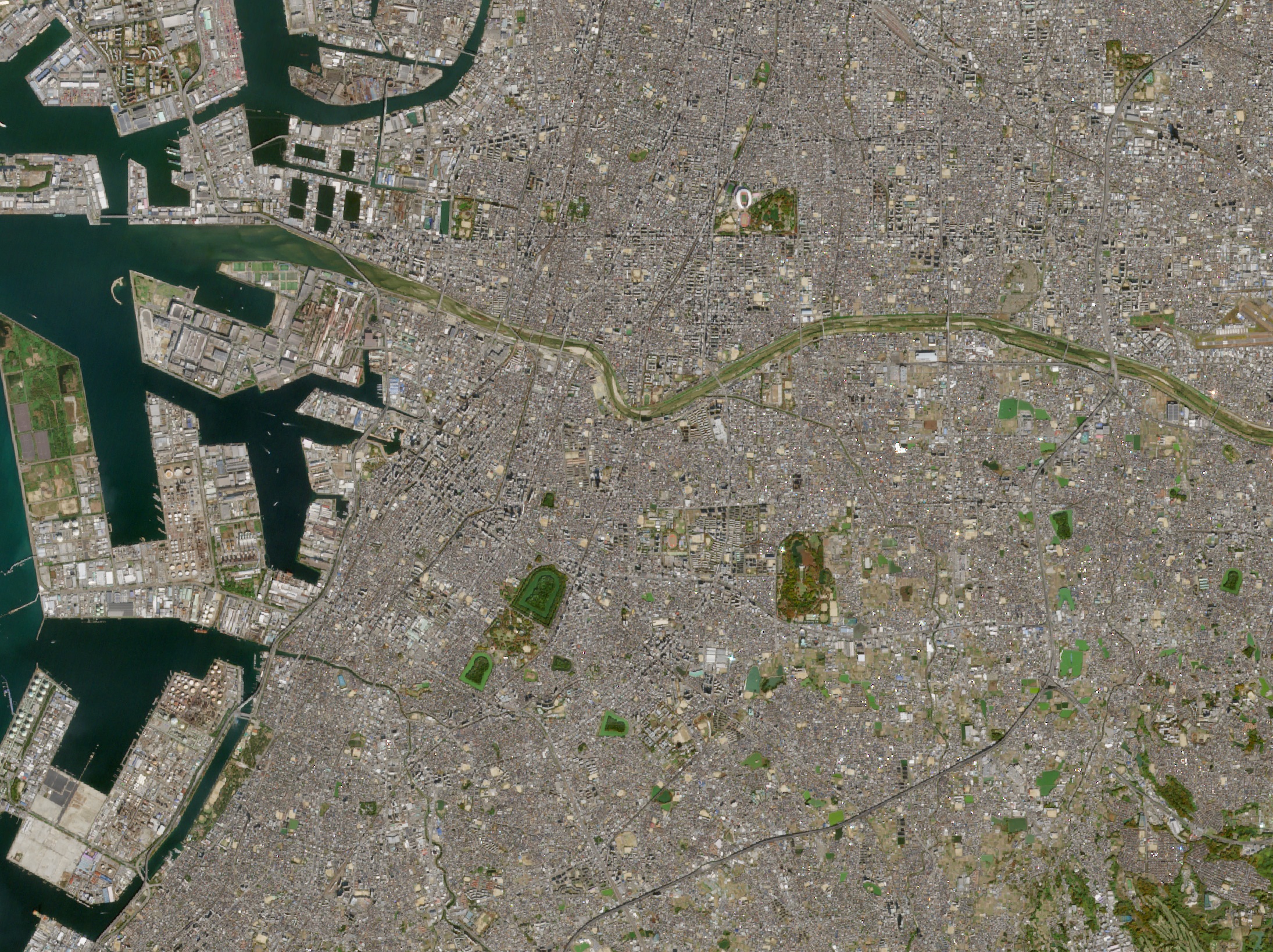}}
	~
	\subfloat[\label{subfig:optical_cloudy}]{%
		\includegraphics[width=0.45\linewidth]{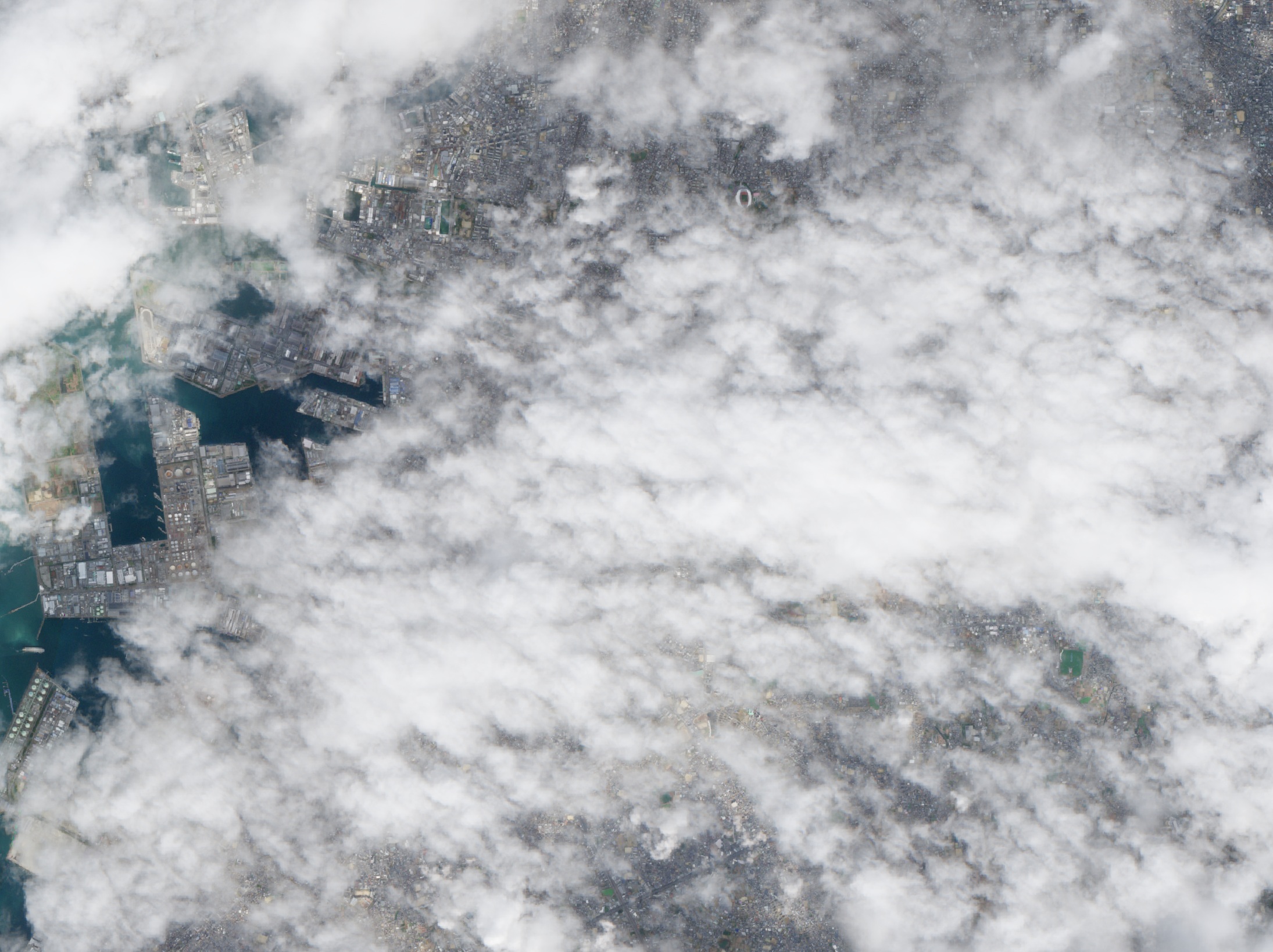}} \\
	\subfloat[\label{subfig:radar_cloudfree}]{%
		\includegraphics[width=0.45\linewidth]{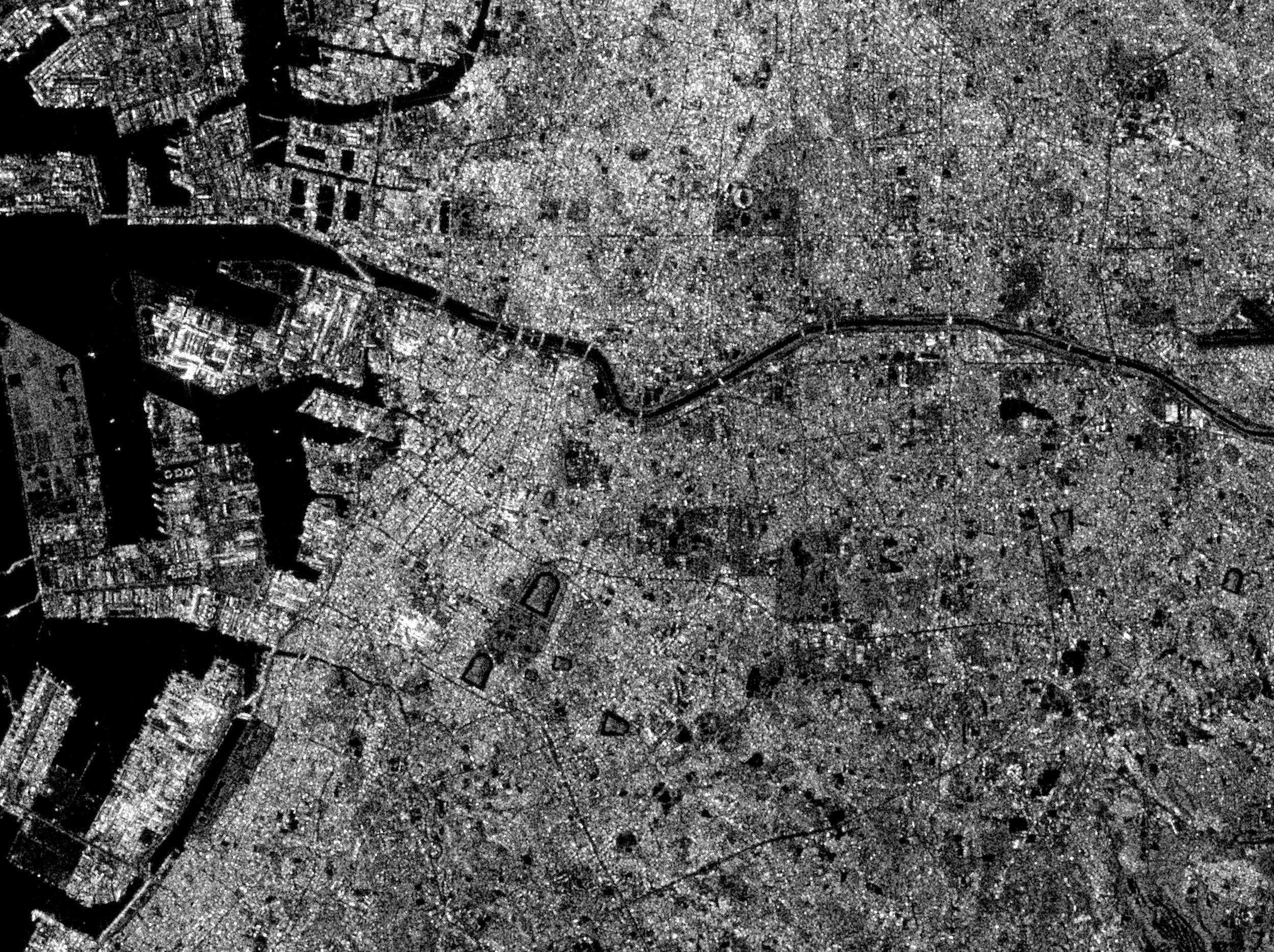}}
	~
	\subfloat[\label{subfig:radar_cloudy}]{%
		\includegraphics[width=0.45\linewidth]{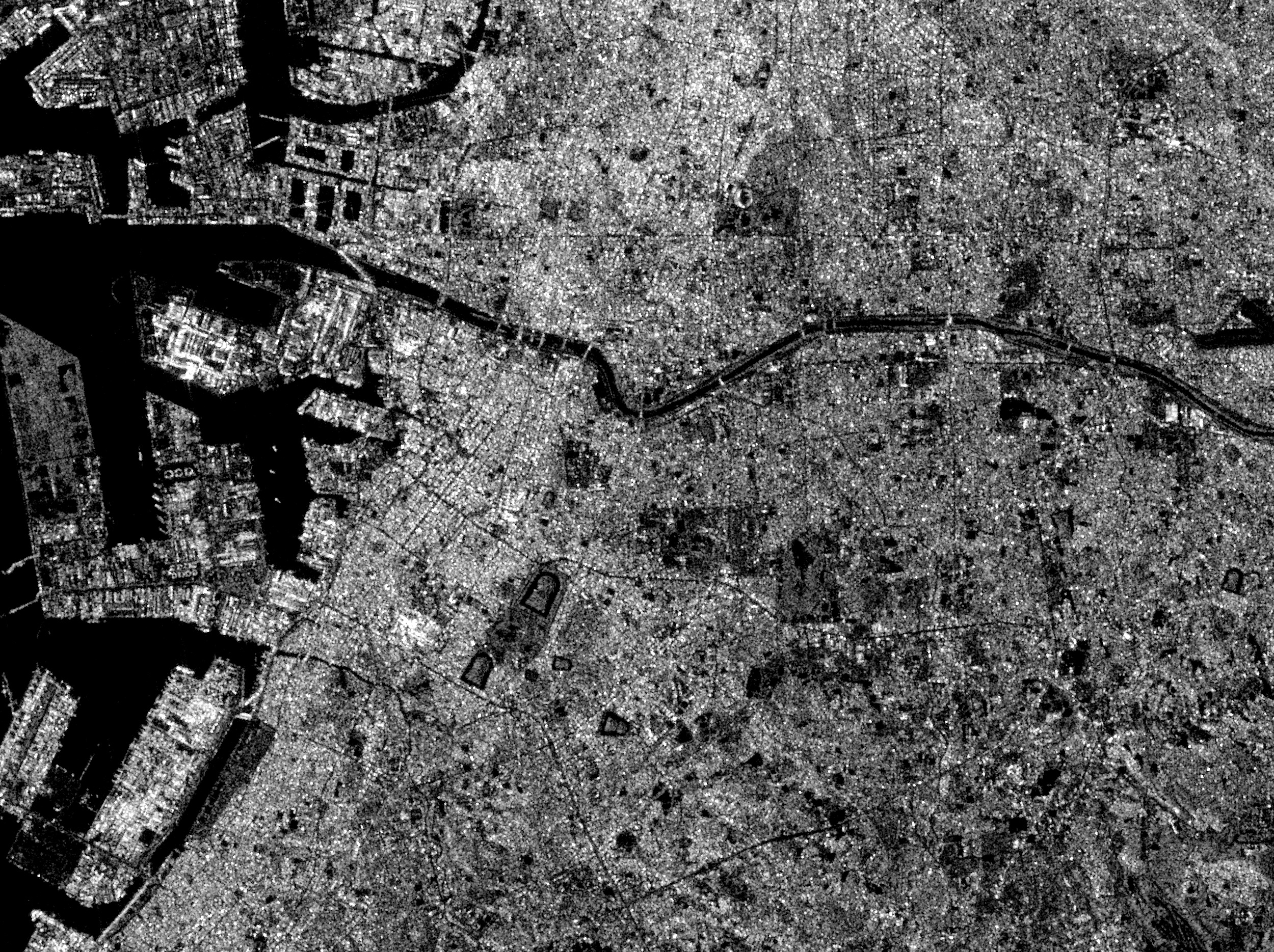}}
	\caption{Top row shows optical (RGB channels) of a subarea of the prefecture of Osaka during a cloud free and a cloudy time period. Bottom row shows radar images (VH channel) at a similar date compared to the corresponding image above.}
	\label{fig:optic_radar} 
\end{figure}

In contrast, cloud robust analysis aims at overcoming cloudy imagery by including cloud piercing technologies, such as radar satellites, as an additional data modality. Radar sensors actively emit signals and record their backscatter. This produces images that are harder to interpret than optical images, but instead are unaffected by clouds. An example optical and radar image pair is shown in Figure \ref{fig:optic_radar}. Such complementary satellite images are readily available from constellations such as Sentinel which produces several terabytes of data per day\cite{sudmanns2020assessing}, offering a rich corpus from which machine learning based methods can automatically extract valuable insights. On one hand  machine learning datasets and applications are using radar and optical modality in order to improve the performance compared to uni-modal models\cite{wang2023ssl4eo,bonafilia2020sen1floods11,helber2019eurosat,sumbul2021bigearthnet,velazquez2025earthview,li2022mcanet,rambour2020flood,bai2025orbit,guo2024skysense,astruc2024omnisat,wang2022self}. However, on the other hand all of these datasets include only cloud free optical images, nor do they investigate the impact of clouds on developed models. Therefore, the area of cloud robust analysis remains largely unexplored. 

To tackle this shortcoming we create a novel dataset, CBEN -- CloudyBigEarthNet, which contains cloudy distortions and introduce machine learning methods to it to solve a Land-Use-Land-Classification (LULC) task in a cloud robust manner. Furthermore, we show that, machine learning methods trained on datasets containing optical and radar data but excluding cloud distortions cannot generalize equally well to cloud free and cloudy imagery. Therefore, without CBEN, evaluation and finetuning in cloudy weather conditions is impossible.

To summarize our contributions are:

\begin{itemize}
	\item We create a downstream-task dataset, CBEN -- CloudyBigEarthNet, with cloud distortions for training and evaluating machine-learning methods and publish it here: \url{https://github.com/mstricker13/CBEN}
	\item We evaluate multiple state-of-the-art machine learning methods on both cloud free and cloudy data, using average precision (AP) as the evaluation metric, and show that a significant performance drop can be observed for methods trained on cloud free imagery when applied to cloudy data, with a decrease between 23 and 33 percentage points.
	\item We propose a data-driven finetuning approach that is robust to both cloud free and cloudy conditions; this approach outperforms the baseline machine learning methods on our cloudy test cases by between 17.23 and 28.72 percentage points. 
\end{itemize}

The paper in hand is structured as follows: Section \ref{sec:rw} briefly explains the related work. This is followed by a detailed description of datasets we have utilized and created in Section \ref{sec:data}. The methods and how they are used with this data are illustrated in Section \ref{sec:method}. Our results and discussion of them are given in Section \ref{sec:res}. Lastly, Section \ref{sec:conclusion} gives our concluding remarks.   

\section{Related Work}
\label{sec:rw}

Cloud robust analysis for machine learning is a largely unexplored topic in earth observation, even though it is highly relevant. In this section we briefly summarize related work most relevant to our study: machine learning datasets for remote sensing, cloudless machine learning, cloud removal, and cloud robust machine learning. 

\subsection{Datasets in Remote Sensing}
A multitude of datasets exist for various applications. Popular datasets for land-use/land-cover classification are BigEarthNet with a cloud coverage less than 1\%, EuroSAT or FMoW, both with with a low cloud percentage as described by the respective authors \cite{helber2019eurosat,sumbul2021bigearthnet,christie2018functional}. DeepGlobe 2018 extends land-use/land-cover classification with road extraction and building detection for optical only modality, where cloudy pixels make up 0.04\% of all pixels \cite{demir2018deepglobe}. Flood detection is covered by Sen1Floods11 which has discarded cloudy images as well \cite{bonafilia2020sen1floods11}. An extensive overview of many different applications is given by et Schmitt al. \cite{schmitt2023there}.

However, generating labels for large-scale earth observation data is time consuming and often requires expert knowledge, even though this domain produces large amounts of unlabeled data-points regularly\cite{sudmanns2020assessing}. Self-supervised learning (SSL) addresses this limitation by exploiting unlabeled data to pretrain powerful feature extractors that are adapted to specific use cases, or downstream tasks \cite{wang2022self,bai2025orbit}. To facilitate SSL several dedicated datasets for remote sensing have been published. Stewart et al. introduced an SSL dataset built from a large historic Landsat time series that contains only optical images where each image has less than 20\% cloud coverage \cite{stewart2023ssl4eo}. Wang et al. created a Sentinel-based dataset that combines radar and optical imagery, with less than 10\% cloud coverage, for SSL \cite{wang2023ssl4eo}. Most recently, Velazquez et al. published EarthView for SSL, which contains optical, radar and lidar data of varying spatial resolution with a cloud coverage less than 30\%, acquired from Satellogic, Sentinel and NEON \cite{velazquez2025earthview}. To the best of our knowledge, these are the only publicly available datasets that target EO-specific SSL. Importantly, all three datasets deliberately avoid images with substantial cloud cover. Private SSL datasets exist, such as the dataset utilized by SkySense to achieve impressive results \cite{guo2024skysense,zhang2025skysense}. Due to the unavailability we do not consider such datasets in our experiments. Based on large pretraining datasets, SSL methods have achieved state-of-the-art results in tasks including semantic segmentation, object detection, change detection and scene classification \cite{guo2024skysense,astruc2024omnisat}.

\subsection{Cloudless Machine Learning}
Based on large pretraining datasets, SSL methods have achieved state-of-the-art results in tasks including semantic segmentation, object detection, change detection and scene classification \cite{guo2024skysense,astruc2024omnisat,bai2025orbit,wang2022self}. Many successful SSL methods originally developed for natural images have been adapted to remote sensing. For example, Wang et al. evaluate both generative approaches (masked autoencoders) and contrastive approaches (MoCo) on their SSL4EO dataset \cite{wang2023ssl4eo,he2022masked,chen2020improved}. These works commonly use ResNet and Vision Transformer (ViT) backbones \cite{wang2023ssl4eo,he2016deep,dosovitskiy2020image}. Because we are utilizing these approaches in our experiments, we will explain them in greater detail in Section \ref{sec:exp}. Beyond direct adaptations, several papers propose EO-specific SSL designs: Wang et al. introduce inter- and intra-model embedding mechanisms to better align modalities during pretraining \cite{wang2024decoupling}. Fuller et al. combine contrastive and reconstructive objectives to learn unimodal and multimodal representations separately \cite{fuller2023croma}. Astruc et al. follow an Omnivore-inspired pipeline that encodes patches of very-high-resolution imagery, radar time series, and optical time series into a unified multimodal representation which is then decoded into each modality \cite{girdhar2022omnivore,astruc2024omnisat}. 

As these works demonstrate, SSL is currently among the best performing paradigms for many EO tasks because it leverages large volumes of unlabeled data to mitigate the scarcity of labeled examples. However, as noted in Section \ref{sec:intro}, all of these methods are trained and evaluated on cloud free collections, whereas real-world satellite imagery contains a large fraction of cloudy observations. 

\subsection{Cloud Removal}
Cloud removal methods are one of the few areas which specifically include cloudy images in their data for machine learning. Lin et al. published two datasets RICE-I, based on Google Earth, and RICE-II, based on Landsat-8\cite{lin2019remote}. Each of them contains no corresponding SAR data. Similar, STGAN and WHUS2-CR don't contain SAR data, but instead replace Landsat-8 with Sentinel-2 to achieve a higher spatial resolution\cite{sarukkai2020cloud,li2021deep}. SAR satellite data is included in SEN12MS-CR\cite{ebel2022sen12ms} and SEN12MS-CR-TS\cite{ebel2023uncrtaints}, where the latter contains a time series of images per location. Most recently, M3R-CR, has been published which contains optical and radar data\cite{xu2023multimodal}. Due to the inclusion of Planetscope data, the spatial resolution is increased further by M3R-CR.

To tackle cloud removal many modern approaches are making use of deep learning\cite{xu2023multimodal}. For example Enomoto et al. and et al. are both utilizing generative adversarial networks as their base method\cite{enomoto2017filmy,singh2018cloud}. Several works, are considering time series information, or most recent cloud free images as sources for inpainting methods\cite{sarukkai2020cloud,czerkawski2022deep}. However, inpainting methods are less effective for regions with persistent cloud cover\cite{xu2023multimodal}. A state-of-the-art approach by Ebel et al., uses an attention mechanism and provides multivariate uncertainty estimates for the reconstructed pixels \cite{ebel2023uncrtaints}. 

Nevertheless, as mentioned in Section \ref{sec:intro}, cloud removal faces several challenges. These include the pairing of input data with ground truth data\cite{zhang2024lightweight}. Creating artificial clouds instead, via for example Perlin noise or extraction of clouds from images, do not correlate to natural cloud occurrences\cite{ebel2022sen12ms}. Furthermore, difficult edge cases caused by cloud shadows or snowy regions can lead to errors. Lastly, reconstructed regions may exhibit visible artifacts or altered brightness relative to their surroundings, a problem that increases with a higher amount of cloud coverage \cite{ma2023cloud}.

\subsection{Cloud Robust Analysis}
Recently, few works have been published with the goal of providing cloud robust analysis. Ling et al. applies data fusion to generate a SAR-optical dictionary space\cite{ling2024enhancing}. However, their approach is only applied on 5 locations within China, showing a geographical and quantitative limitation. Furthermore, no comparison to other state-of-the-art machine learning models has been done. The same issues are encountered in the work of Prudente et al., which only covers a single area in Brazil\cite{prudente2022multisensor}. Most closely, related to our work is CloudSeg by Xu et al\cite{xu2024cloudseg}. CloudSeg trains a network with radar and optical data on cloud removal and LULC together in order to achieve robust results for LULC on images contaminated by clouds. However, their dataset has been created with artificially generating clouds via Perlin noise. This approach doesn't correlate to real cloud statistics in remote sensing images, as pointed out by Ebel et al. and therefore can still fail in real world applications\cite{ebel2022sen12ms}. The datasource used by Xu et al. provides 4 optical bands (Red, Green, Blue, Near Infra-red), whereas Sentinel-2, as utilized by us, provides 13. Other research shows that performance gains are observed with more spectral bands\cite{jocea2025sentinel}. Because, depending on the task, other bands can include critical information\cite{wang2023mapping,delegido2011evaluation}.

Because perfect cloud removal remains an open problem and many state-of-the-art application models are SSL-based and trained on cloud free data, it is crucial to evaluate how these established methods perform under realistic, cloudy conditions, giving further critical insights toward this challenge. To the best of our knowledge, this paper is the first systematic study that investigates the impact of clouds on SSL-based methods. The importance of such a study is also pointed out by Schmitt et al. survey paper on datasets in remote sensing \cite{schmitt2023there}.

\section{Data}\label{sec:data}
The section describes the remote-sensing modalities used in this work, the datasets employed for self-supervised pre-training, downstream task evaluation and the modifications we applied to create a realistically cloudy variant of BigEarthNet. 

\subsection{Remote Sensing Modalities}\label{subsec:rs}
We focus on the two complementary modalities provided by the Sentinel mission: optical imagery from Sentinel-2 and radar imagery from Sentinel-1. Here, we highlight their respective advantages and disadvantages in the context of this paper. 

Optical sensors record sunlight reflected from the Earth's surface across multiple wavelengths. Sentinel-2 delivers multi-spectral observations in 13 bands, or otherwise known as channels, covering 443 nm to 2190 nm \cite{spoto2012overview}. Several of these bands correspond to semantic meaningful wavelengths. These include the visible red, green and blue wavelengths, which makes optical images relatively intuitive to interpret even for non-experts. Optical imagery is rich in semantic content and typically provides the most expressive features for many machine-learning tasks \cite{wang2023ssl4eo,shen2023evaluating,he2025optical,crowson2019comparison}.

Synthetic-aperture radar (SAR) sensors actively emit microwave signals and measure the backscatter returned from the surface. SAR imagery is largely insensitive to illumination and cloud cover, enabling nighttime and through-cloud observations. SAR polarimetric channels encode the transmit/receive signal orientation combinations, for example VV, VH, where ``V'' denotes vertical and ``H'' denotes horizontal polarization \cite{yun2019research}. SAR reflects the physical surface but is generally less visually intuitive than optical bands. Thus, optical and SAR modalities are complementary. Optical data carry highly discriminative spectral cues, while SAR provides reliable structural information under adverse weather.

Many existing datasets and benchmarks deliberately exclude images with substantial cloud cover, as mentioned in Section \ref{sec:rw}. This choice is sensible in specific contexts. 

First, manual labelling is difficult or impossible when labels must be inferred beneath opaque clouds, as annotating SAR data typically requires domain expertise whereas optical RGB facilitates annotation by non-experts. 

Secondly, when training only on optical data, cloud free images prevent learning erroneous features. Because the clouds don't correspond to the ground truth label information unlike the underlying surface which produces the discriminative features relevant for the labels.

Lastly, optical data contains the most relevant features for machine learning methods. Therefore, cloud free data is beneficial to achieve the best performing results, as this condition prevents disadvantageous distortions in this modality, which would reduce the information content for predictions. Ablation studies in the literature have analyzed multi-modal models combining optical and radar data against uni-modal models, which make either use of optical or radar data separately. Generally, the best performance is achieved by multi-modal models, with optical only models having a slightly worse performance, but still outperforming radar only models. Indicating, that even for the machine learning model the most expressive features are contained in the optical modality\cite{wang2023ssl4eo,shen2023evaluating,he2025optical,crowson2019comparison}. 

However, excluding cloudy images is suboptimal when both optical and SAR observations are available and when methods are intended for realistic and operational settings or time-sensitive applications such as disaster response \cite{schmitt2023there}. Clouds are not preventing feature extraction when both modalities are available. Then, the technological solution, SAR imagery can be utilized to pierce clouds and extract information from the underlying ground. Furthermore, in self-supervised learning there is no requirement for human annotation, so cloudy imagery can be exploited directly and automatically for pretraining. This enables the development of EO-specific SSL objectives and multi-modal strategies that leverage SAR to compensate for optical occlusion. These strategies can be unique for EO and distinct from the approaches defined for natural images.  

\subsection{SSL4EO-S12}\label{subsec:datassl4eo}
For pretraining we use the SSL4EO-S12 dataset \cite{wang2023ssl4eo}. SSL4EO-S12 pairs Sentinel-1 SAR and Sentinel-2 optical observations from the same location at approximately the same acquisition time. Optical data in the collection are provided as Level-1 (top-of-atmosphere) and Level-2 processing levels, where Level-2 products have had atmospheric effects removed. 

\begin{figure}
	\centering
	\includegraphics[width=\linewidth]{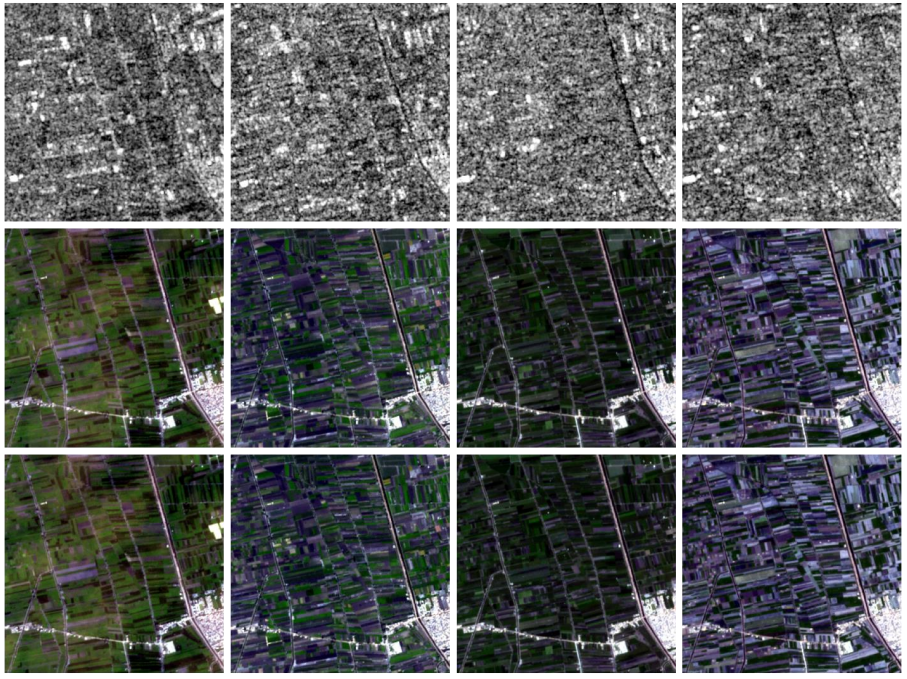}
	\caption{Example location of SSL4EO-S12. The columns from left to right represent the timestamps at spring, summer, fall and winter. The rows from top to bottom represent radar, and optical level 1 and level 2 data.}
	\label{fig:ssl4eosample}
\end{figure}

SSL4EO-S12 contains diverse data covering all seasons and regions. Each sampled location in SSL4EO-S12 includes four temporally distinct observation pairs corresponding to the four meteorological seasons (winter, spring, summer, autumn), thereby capturing seasonal variability at each site. The dataset is globally distributed to cover diverse regional characteristics. A representative example showing the SAR and Level-1/Level-2 optical modalities for a single location across seasons is given in Figure \ref{fig:ssl4eosample}.

We note that shortly after completing our experiments SSL4EO-S12 was updated to correct minor geographic misalignments between Sentinel-1 and Sentinel-2 samples \cite{blumenstiel2025ssl4eo}. The versions used in our study has, however, been employed in several prior works that report state-of-the-art results \cite{fuller2023croma,astruc2024omnisat,astruc2025anysat}. Repeating the full SSL pretraining with the updated release was not feasible within our experimental budget, and no comprehensive performance comparison between the two releases has been reported \cite{blumenstiel2025ssl4eo}.

\subsection{BigEarthNet}\label{subsec:BEN}
For donwstream evaluation we use BigEarthNet \cite{sumbul2021bigearthnet}, as it is also being utilized in several other publications \cite{fuller2023croma,astruc2024omnisat,nedungadi2024mmearth}. 

BigEarthNet contains 549,488 paired Sentinel-1 (SAR) and Sentinel-2 (optical) samples with multi-label annotations for LULC classification. The labels are derived from the CORINE Land Cover (CLC) taxonomy, and the publicly released dataset provides two label granularities: 43 fine-grained classes and a coarser 19 class grouping. In this work we use the 19 class variant. The labels are given in Table \ref{tab:BEN_Labels}. 

\begin{table}[]
	\caption{Tabular listing of all 19 target labels contained in BigEarthNet\label{tab:BEN_Labels}}
	\begin{tabular}{@{}l@{}}
		\toprule
		Label                                                                                  \\ \midrule
		Agro-forestry area                                                                     \\
		Arable land                                                                            \\
		Beaches, dunes, sands                                                                  \\
		Broad-leaved forest                                                                    \\
		Coastal wetlands                                                                       \\
		Complex cultivation patterns                                                           \\
		Coniferous forest                                                                      \\
		Industrial or commercial units                                                         \\
		Inland waters                                                                          \\
		Land principally occupied by agriculture                                               \\
		\quad with significant areas of natural vegetation                                     \\
		Marine waters                                                                          \\
		Mixed forest                                                                           \\
		Moors, healthland and sclerophyllous vegetation                                        \\
		Natural grassland and sparsely vegetated areas                                         \\
		Pastures                                                                               \\
		Permament crops                                                                        \\
		Transitional woodland, shrub                                                           \\
		Urban fabric                                                                           \\ \bottomrule
	\end{tabular}
\end{table}

\begin{figure} 
	\centering
	\includegraphics[width=\linewidth]{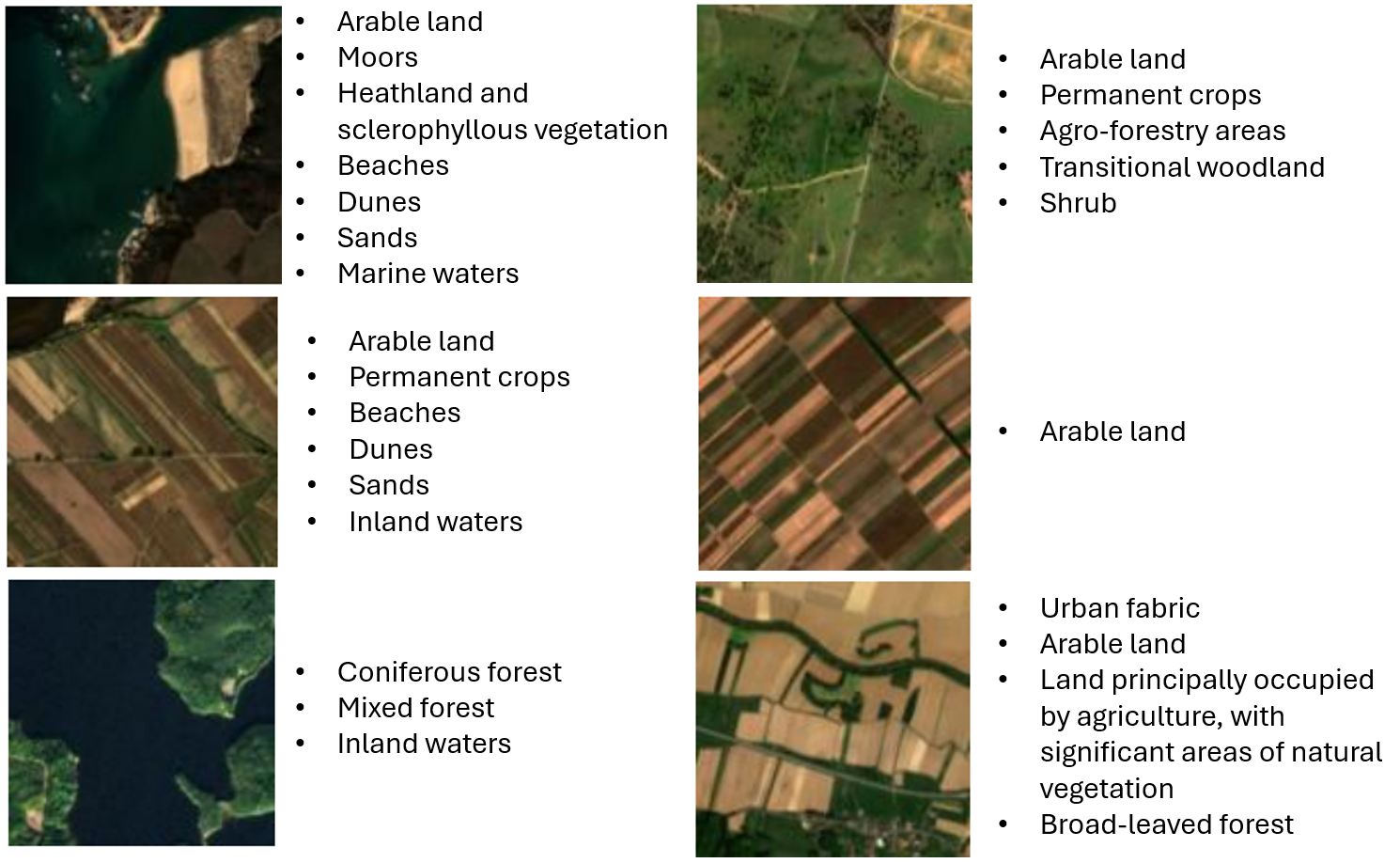}
	\caption{Visualization of several samples of BigEarthNet with corresponding labels.}
	\label{fig:ben_samples} 
\end{figure}

BigEarthNet was constructed from largely cloud free Sentinel imagery over Europe with a reported cloud coverage of less than 1\%. Labels are assigned at the image (tile) level. Each sample may have one or more class labels describing the dominant land-use/land-cover types in the tile. Because the label assignment is global to the tile rather than pixel wise, BigEarthNet targets classification tasks rather than semantic segmentation. The catalogue of classes covers persistent land-cover categories (e.g., broad-leaved forest, urban fabric, arable land) that typically change slowly over time. We therefore, refer to these as \textit{static} labels. Small scale and dynamically changing labels, such as moving objects, e.g. a plane, are not covered in this dataset. Example images showing the optical modality with the corresponding labels are given in Figure \ref{fig:ben_samples}.

\subsection{CBEN -- CloudyBigEarthNet}\label{subsec:CBEN}
In this section we describe the construction of a cloudy variant of BigEarthNet, see Section \ref{subsec:BEN}, to study the robustness of SSL-trained models under realistic cloud conditions. The adaptation leverages two properties of BigEarthNet: (i) its labels correspond to largely \textit{static}, coarse land-cover categories and (ii) each tile is associated with precise spatial and temporal metadata (tile identifier, location, acquisition date).

The central assumption underlying our conversion is that \textit{static} tile level labels remain valid across short temporal differences. That is, although local pixel level appearance may change (e.g. shadows, seasonal vegetation, transient flooding), the dominant land-cover label for a tile generally persists over short time scales. This assumption is commonly employed in remote-sensing tasks that align optical and SAR observations \cite{sumbul2021bigearthnet,wang2023ssl4eo} or reuse labels across temporally adjacent acquisitions, such as it is done for cloud removal tasks \cite{ebel2020multisensor,ebel2022sen12ms,schmitt2019sen12ms,schmitt2019aggregating}. However, cloud removal is defined by a pixel level reconstruction task, thereby being affected by local pixel changes through time, whereas the general tile level information, as in our case, stays the same. Bastani et al. utilizes a similar idea to train a foundational model, on cloud free data, that uses input images at different time-steps with only one target label image \cite{bastani2023satlaspretrain}. Lastly, Ling et al. applies a similar idea to evaluate cloudy images\cite{ling2024enhancing}. However, no large scale dataset for machine learning has been created.

\begin{figure}
	\centering
	\includegraphics[width=\linewidth]{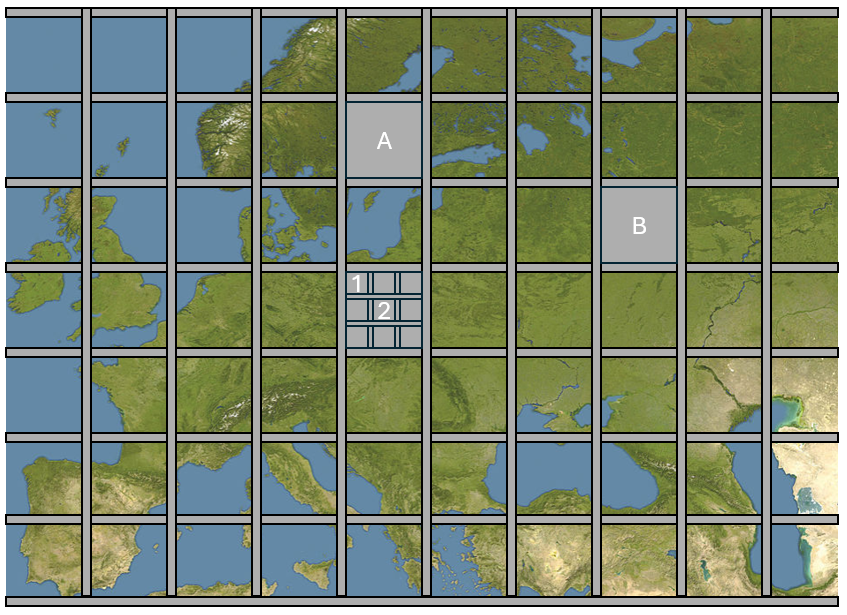}
	\caption{Illustration of how BigEarthNet downloaded images. Boxes with literals are called tiles. Boxes with numericals are named patches.}
	\label{fig:BEN_tile_patch}
\end{figure}

To create our dataset we follow the image creation of tiling and dividing the tiles of BigEarthNet. The procedure is to partition Europe into large spatial tiles, which are further subdivided into smaller image patches. This tiling including the naming convention is illustrated in Figure \ref{fig:BEN_tile_patch}. BigEarthNet excludes tiles with substantial cloud cover larger than 1\% \cite{sumbul2021bigearthnet}. 

Utilizing the locational and temporal information of each tile, we access the publicly available cloud cover dataset created by Wilson et al.\cite{wilson2016remotely}. Wilson et al. provides for any land based location, over the globe, the mean probability and standard deviation of a cloud occurring there. This statistics are given on a per-month basis and were derived from multi-year observations. Therefore, they provide a realistic estimate of typical cloudiness at a given place and time. 

\begin{figure}
	\centering
	\subfloat[\label{subfig:large_cloud_tile}]{%
		\includegraphics[width=0.4\linewidth]{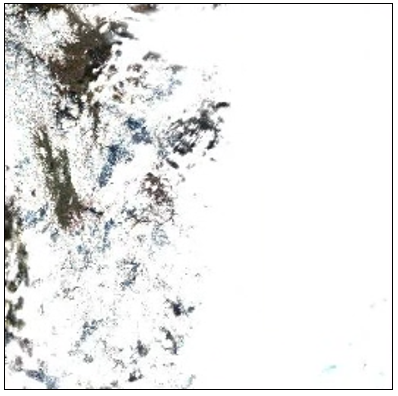}}\qquad
	\subfloat[\label{subfig:small_cloud_tile}]{%
		\includegraphics[width=0.4\linewidth]{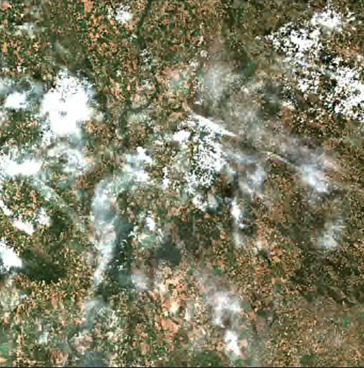}}\\
	
	\subfloat[\label{subfig:large_cloud_patch}]{%
		\begin{minipage}{0.4\linewidth}
			\centering
			\includegraphics[width=0.5\linewidth]{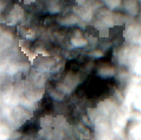}
	\end{minipage}}\qquad
	\subfloat[\label{subfig:small_cloud_patch}]{%
		\begin{minipage}{0.4\linewidth}
			\centering
			\includegraphics[width=0.5\linewidth]{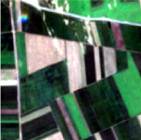}
	\end{minipage}}
	
	\caption{Visualization of different tiles and patches of our dataset CBEN with varying cloud cover percentages. (a) Tile with large cloud cover percentage, (b) Tile with low cloud cover percentage, (c) Patch with large cloud cover percentage, extracted from respective tile, (d) Patch with low cloud cover percentage, extracted from respective tile.}
	\label{fig:cloudcoverdata}
\end{figure}

The cloud cover information is then used to download tiles, covering the same locations as in BigEarthNet, which are temporally as close as possible to the respective reference tile while satisfying our cloud cover percentage. Using the location and timestamp of a BigEarthNet reference tile, we request from the ESA archive the temporal closest Sentinel-2 acquisition whose cloud fraction is approximately equal to the queried mean probability plus/minus the standard deviation. From the downloaded Sentinel-2 tile we extract patches using the same tiling scheme employed by BigEarthNet, thus producing image patches whose cloud occurrence rates reflect realistic, location- and season dependent statistics. Across our constructed dataset, the mean cloud coverage is 65.32\% with a standard deviation of 16.56\%. These values are summarized in Table \ref{tab:cloud_cover}. Figure \ref{fig:cloudcoverdata} shows two example tiles and a corresponding patch from our cloudy dataset. This figure highlights that we have downloaded images with large cloud cover, creating mostly cloudy patches, but also images with small cloud cover, creating almost cloud free patches as well.

\begin{table}[]
	\centering
	\caption{Statistics of cloud related distortions in the downloaded tiles. Values have been extracted from the metadata provided by ESA.}
	\begin{tabular}{@{}llll@{}}
		\toprule
		& Cloud Cover & Cloud Shadows & Thin Cirrus Clouds \\
		\midrule
		Mean               & 65.32       & 2.98          & 11.91              \\
		Standard Deviation & 16.56       & 3.37          & 14.33              \\
		Minimum            & 17.03       & 0.00          & 0.00               \\
		Maximum            & 98.98       & 18.26         & 55.07             
		\label{tab:cloud_cover}
	\end{tabular}
\end{table}

For SAR we reuse the Sentinel-1 images originally provided by BigEarthNet \cite{sumbul2021bigearthnet}. To quantify temporal alignment we compute the mean temporal offset between the original BigEarthNet optical images and their corresponding SAR images, the temporal distance between BigEarthNet optical images and our newly downloaded cloudy optical images, and the temporal distance between the newly downloaded optical images and the reused SAR images. The statistics of these temporal distances are reported in Table \ref{tab:temp_dist}. These statistics inform the expected temporal mismatch introduced by substituting the original cloud free optical observations with temporally adjacent, realistically cloudy acquisitions. 

\begin{table}[]
	\centering
	\caption{Temporal distances between BigEarthNet S2 images to S1 images and our S2 cloudy images to BigEarthNet's S1 and S2 images.}
	\begin{tabular}{@{}llll@{}}
		\toprule
		& S2 to S1        & S2 Cloud to S1    & S2 Cloud to S2    \\
		\midrule
		Mean               & 0 days 20:50:06 & 9 days 07:51:51  & 9 days 06:42:41  \\
		\makecell{Standard \\ Deviation} & 0 days 18:05:07 & 8 days 04:29:50  & 8 days 04:47:11  \\
		Minimum            & 0 days 04:40:41 & 0 days 04:53:03  & 1 days 00:19:48  \\
		Maximum            & 5 days 17:48:54 & 42 days 06:14:20 & 40 days 00:00:00
		\label{tab:temp_dist}
	\end{tabular}
\end{table}

\section{Methods}\label{sec:method}
This section describes the self-supervised learning (SSL) techniques and network architectures used in our experiments. We motivate our design choices and provide concise technical descriptions of each method. 

\subsection{Self-Supervised Learning}\label{subsec:ssl}
Self-supervised learning leverages large corpora of unlabeled data to pretrain feature extractors (backbones) using automatically generated labels, which are then adapted to downstream tasks by training a lightweight prediction head. SSL is particularly attractive for Earth observation because of the abundance of unlabeled satellite imagery. As described in Section \ref{sec:rw}, SSL based methods have shown to achieve state-of-the-art performances due to this reason.

Supervised machine learning is being trained by using ground truth information, the labels, which are annotated by humans. The key idea behind supervised machine learning methods, is that first the programmer defines the structure of a  mathematical model, such as a neural network. Then, the parameters of this model are being trained using input data and corresponding ground truth labels \cite{schmidhuber2015deep}. The capabilities of such a model correlate to the size of the dataset used. With more data, more complex models can be trained, that in turn can learn more complex underlying structures and therefore perform better \cite{sun2017revisiting}. Creating such large datasets with the corresponding labels is time consuming and costly and cannot be done for every single task. 

To tackle this problem, two directions are investigated, reducing the number of required samples without sacrificing performance and improving performance by utilizing unlabeled data. The first idea proposes methods, such as feature space engineering \cite{stricker2023fusing}, dimensionality reduction \cite{jia2022feature}, or transfer learning \cite{iman2023review}. The second direction includes approaches such as weakly supervised learning\cite{zhou2018brief}, semi-supervised \cite{yang2022survey} learning or self-supervised learning \cite{rani2023self}. Nevertheless, self-supervised learning has emerged as a promising solution to this problem as it can make use of a large corpus of unlabeled data and can then be applied on many applications \cite{rani2023self}. 

The key idea of SSL is that first a large and complex neural network is trained on a big corpus of unlabeled data by applying certain modification to the input data. As the modifications were done in an algorithmic procedure, the parameters of this modification are used as the ground truth labels. Predicting the modifications is known as the self-supervised task. The network trained in this way, will then be able to extract well defined features from the input data, and is defined as backbone. Then, the last layers of the network, responsible for predicting the labels, are called head. Similar to transfer learning, by reusing the backbone and retraining the head on another small scale dataset the network can be adapted to different tasks that make use of similar input modalities. The new task, the network is adapted to is called, downstream task. 

We evaluate three representative SSL approaches that span reconstructive and contrastive paradigms: Masked Autoencoders (MAE) \cite{he2022masked}, MoCo v2 \cite{chen2020improved} and MoCo v3 \cite{chen2021empirical}. MAE is included because its masking objective (predicting occluded image regions) is conceptually similar to the occlusion induced by clouds and may therefore impart robustness to missing pixels. MoCo variants provide complementary contrastive learning baselines, to ensure that our conclusions do not depend on a single SSL method but the lack of cloudy imagery. 

\subsubsection{Masked Autoencoder}
\begin{figure}
	\centering
	\includegraphics[width=\linewidth]{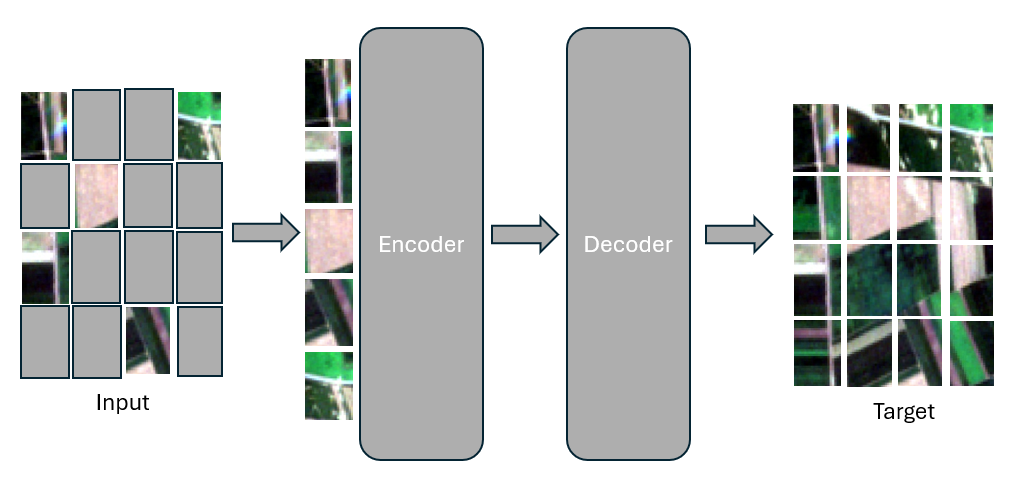}
	\caption{Visual representation of the self-supervised task masked autoencoder}
	\label{fig:mae}
\end{figure}

Masked Autoencoders, as visualized in Figure \ref{fig:mae}, are occluding parts of an image and then recreate the occluded area \cite{he2022masked}. In a first step, the input image is divided into several equal sized patches, where each patch contains multiple pixels. Then, a majority of these patches are masked out. He et al. suggests 75\% as a good value of removed patches \cite{he2022masked}. Only the not masked out patches are fed into an encoder network, which learns and predicts a latent representation based on these few patches. This latent representation is used as an input in addition with the masked out patches and positional information for each patch for the decoder. The task of the decoder is to predict the pixel values of the masked out patches in order to recreate the original image.

\subsubsection{Momentum Contrast}\label{subsec:moco}
Momentum Contrast (MoCo) \cite{he2020momentum} is a contrastive self-supervised learning paradigm. In this paper, we utilize MoCo2, as proposed by Chen et al. \cite{chen2020improved} which includes ideas from SimCLR \cite{chen2020simple}. The key idea is to train an encoder which produces similar representations for augmented views of the same image (positive pairs) and dissimilar representations for different images (negative pairs). MoCo maintains a dictionary (queue) of negative samples and employs a momentum updated encoder to stabilize target representations which improves the consistency between the currently encoded samples and previously encoded samples. MoCo v2 builds on the original MoCo by incorporating stronger data augmentations, larger batch sizes and a multi-layer projection head (MLP) \cite{chen2020improved}.

MoCo v3 adapts contrastive ideas from ResNet \cite{he2016deep} to Vision Transformers (ViTs) \cite{dosovitskiy2020image} and large-scale training \cite{chen2021empirical}. Notable changes include the removal of the explicit negative-sample queue and instead relying on large batch sizes, optimizer and hyperparameter changes, such as utilizing AdamW \cite{loshchilov2017decoupled} as the optimizer or replacing the loss function, and architecture specific adjustments that exploit ViT properties.

To form positive pairs for contrastive learning we apply the same suite of augmentations, inspired by BYOL \cite{grill2020bootstrap}, for both MoCo2 and Moco3 to generate the augmented views. For each view we randomly apply:

\begin{itemize}
	\item Resize and cropping of an image
	\item Changing the brightness and contrast
	\item Transforming the image to grayscale
	\item Applying gaussian blur
	\item Flipping the image horizontaly
\end{itemize}

\subsection{Backbone Networks}

The above mentioned self-supervised learning methods have been implemented with Residual Networks \cite{he2016deep} and Vision Transformers \cite{dosovitskiy2020image} as the backbone networks. We have chosen these backbone networks, as they have been widely adopted in state-of-the-art EO and general computer vision literature \cite{wang2023ssl4eo,janga2023review}. Here, we will now explain the key design ideas behind these networks.

Residual Networks, or ResNets \cite{he2016deep}, are deep neural networks augmented with residual (skip) connections that add the input of a residual block to its output, which alleviates optimization difficulties in very deep networks \cite{krizhevsky2012imagenet}. Typical ResNet architectures consist of a stack of convolutional blocks for feature extraction followed by a pooling and classification head \cite{schmidhuber2015deep,fukushima1983neocognitron,lecun1998gradient}. Convolutional filters extract local spatial features by sliding windows across the image and residual connections allow gradients and features to propagate more effectively across layers.

Vision Transformers (ViTs) \cite{dosovitskiy2020image} adapt the transformer architecture \cite{vaswani2017attention} from natural language processing (e.g. BERT \cite{devlin2019bert}) to images by splitting an image into a sequence of fixed-size patches, projecting each patch to a token embedding and adding positional encodings. The token sequence is processed by multi-head self-attention. ViTs have demonstrated strong performance when pretrained on large datasets, as highlighted in Section \ref{sec:rw}. 

Both backbones are used in our SSL pipelines, where we attach lightweight task-specific heads after pretraining and fine-tune on downstream classification using the labeled BigEarthNet datasets described in Section \ref{sec:data}. 

\section{Experiments}
\label{sec:exp}

This section details our experimental protocol, including data fusion, SSL pretraining, downstream fine-tuning and evaluation.

\subsection{Experimental Setup}

We adapt the publicly available SSL4EO-S12 codebase \cite{wang2023ssl4eo} as the foundation for our implementations, extending the published uni-modal pipeline to a multi-modal setting. We perform early fusion of Sentinel-1 (SAR) and Sentinel-2 (optical) data by concatenating modalities into a single input product with 14 bands: the VV and VH polarimetric channels from Sentinel-1 plus all Level-2A Sentinel-2 optical bands. We use only the Level-2A (surface-corrected) optical products, as explained in Section \ref{subsec:datassl4eo}, to match the preprocessing our our downstream dataset (BigEarthNet) \cite{sumbul2021bigearthnet} and to ensure a consistent channel set between pretraining and fine-tuning. 

All SSL pretraining is performed on the cloud filtered SSL4EO-S12 collection \cite{wang2023ssl4eo} (i.e., the training data does not contain clouds). We pretrain three backbone configurations: ResNet-50 \cite{he2016deep} with MoCo v2 \cite{he2020momentum}, ViT \cite{dosovitskiy2020image} with MoCo v3 \cite{chen2021empirical} and ViT \cite{dosovitskiy2020image} with Masked AUtoencoder (MAE) \cite{he2022masked}. Where possible we reuse the original hyperparameter choices from the references implementations, as they have been shown to achieve promising results in a uni-modal setting. Full training details and exact settings are available in our public code repository.

\subsection{Pretraining Details}
The specific pretraining schedules, optimizers and loss functions are detailed in Table \ref{tab:pretrain_details}

\begin{table*}[]
	\caption{Specific pretraining schedules, optimizers and loss functions used during SSL training}
	\begin{tabular}{@{}llllllll@{}}
		\toprule
		SSL Method         & Backbone  & Number of Epochs & Batch Size & Optimization                                                             & Learning Rate & Loss Function & Notes\\ \midrule
		MoCo v2            & ResNet-50 & 100              & 256        & SGD\cite{robbins1951stochastic} & 0.03          & Cross Entropy & -  \\
		MoCo v3            & ViT       & 100              & 256        & AdamW \cite{loshchilov2017decoupled}                     & 0.15          & Cross Entropy & - \\
		Masked Autoencoder & ViT       & 100              & 64         & AdamW\cite{loshchilov2017decoupled}                     &    0.1           &    Multi Label Soft Margin Loss           &  Masking Ratio 70\% \\ \bottomrule
	\end{tabular}
	\label{tab:pretrain_details}
\end{table*}

For MoCo v2 and v3 we form positive pairs by applying the same set of augmentations to produce different views of each sample. Each view is generated by first randomly resizing and cropping the view. Then, a pipeline of image modifications is applied, where each operation is only executed with a given probability. The operations in order, as well as their probabilities are stated in Table \ref{tab:moco_aug_prob}.

\begin{table}[]
	\caption{The applied image augmentations for MoCo SSL pretraining. An operation is only executed with a given probability}
	\centering
	\begin{tabular}{@{}ll@{}}
		\toprule
		Augmentation                      & Probability \\ \midrule
		Brightness and contrast ajustment & 0.8         \\
		Random grayscale conversion       & 0.2         \\
		Gaussian blur                     & 0.5         \\
		Horizontal flip                   & 0.5         \\ \bottomrule
	\end{tabular}
	\label{tab:moco_aug_prob}
\end{table}

These augmentations mirror the ones described in Section \ref{subsec:ssl} and are chosen to be consistent across MoCo variants.

\subsection{Fine-tuning and Evaluation}
\begin{figure} 
	\centering
	\subfloat[\label{subfig:expa}]{%
		\includegraphics[width=\linewidth]{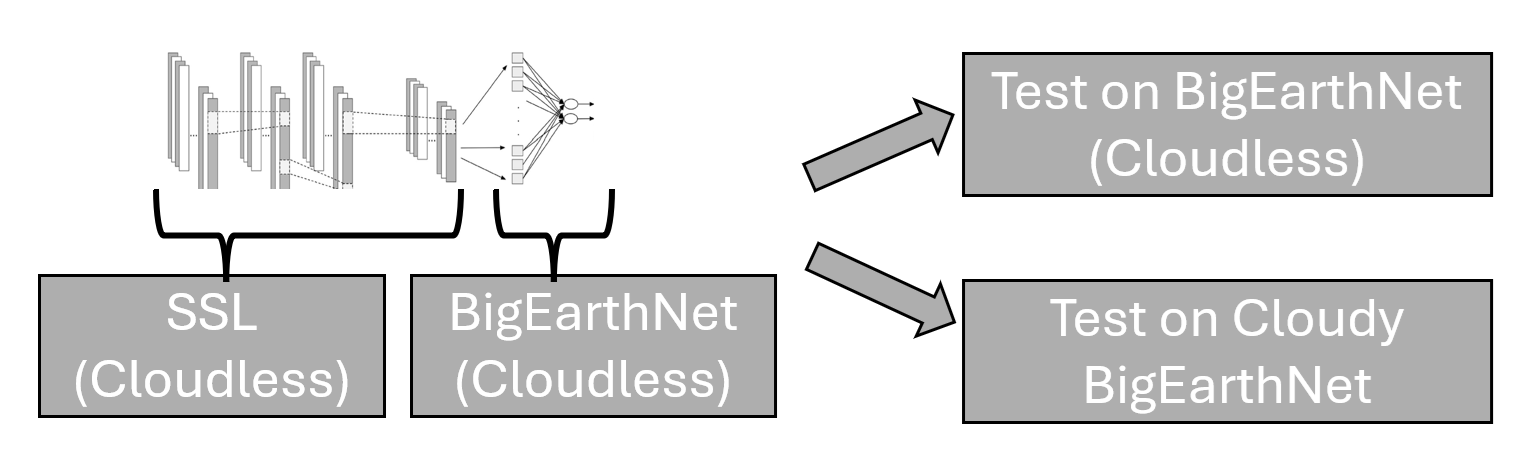}}\\
	\subfloat[\label{subfig:expb}]{%
		\includegraphics[width=\linewidth]{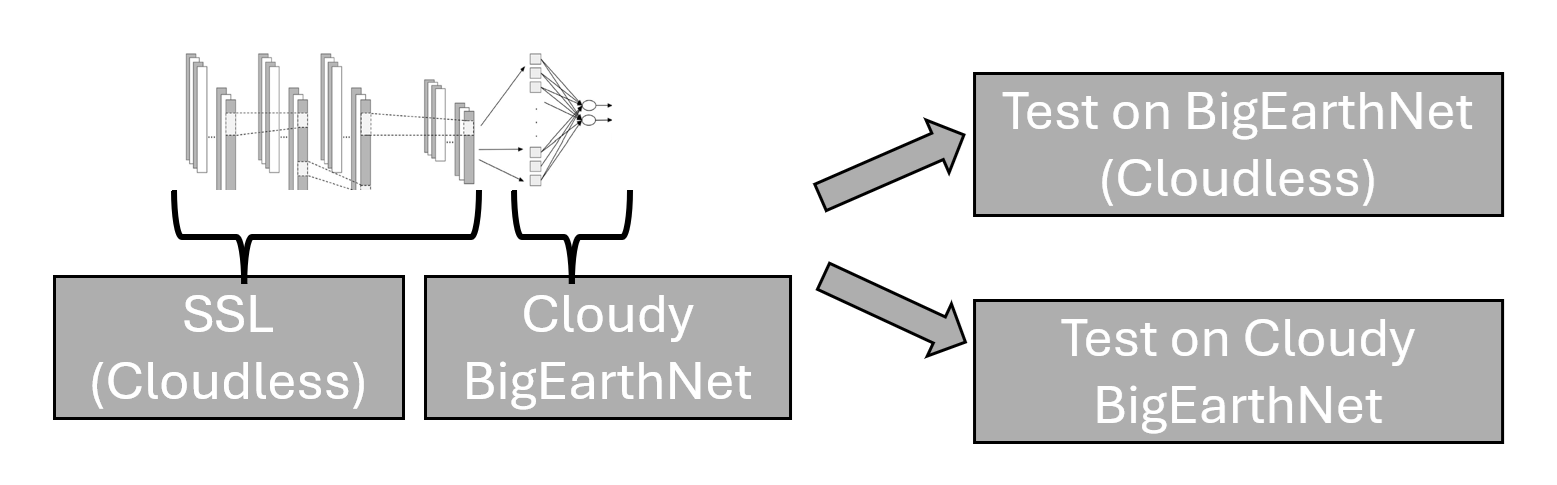}}
	\caption{Visualized experimental architecture. (a) Trained experiment without clouds, (b) Trained experiment with clouds.}
	\label{fig:exp} 
\end{figure}

After SSL pretraining, we discard the pretraining heads and attach classification heads for the downstream LULC task. For each pretrained backbone we train two seperate heads, as illustrated in Figure \ref{fig:exp}:

\begin{enumerate}
	\item a head fine-tuned on the original (cloud filtered) BigEarthNet \cite{sumbul2021bigearthnet} training set
	\item a head fine-tuned on our cloudy BigEarthNet adaptation (see Section \ref{subsec:CBEN})
\end{enumerate}

This procedure yields six trained models (three backbones \texttimes\xspace two finetuning sets). Each of these models is then evaluated on two test sets: the original BigEarthNet test split and our cloudy BigEarthNet test split, yielding a total of 12 experimental evaluations.

We adopt the train/validation/test partitioning proposed by Clasen et al. for BigEarthNet v2.0 \cite{clasen2024reben}. The v2.0 split removes low-quality samples, applies updated ESA atmospheric corrections and uses a location aware splitting strategy. Consequently, direct numerical comparison to results reported on the earlier BigEarthNet release is not possibles, such as the ones performed by Wang et al. during the creation of SSL4EO \cite{wang2023ssl4eo}. 

\subsection{Implementation Notes and Hyperparameter Tuning}
Wherever practical, we retained the original hyperparameters for the downstream task implementation to benefit from published best practices. We did not perform an extensive hyperparameter search due to resource constraints. All hyperparameter choices and training scripts are provided in our released code for reproducibility. 

\subsection{Evaluation Metric}

All experiments are evaluated using the Average Precision ($AP$) as implemented in scikit-learn \cite{pedregosa2011scikit}. $AP$ is computed from the precision-recall curve and represents the weighted mean of precisions at different recall levels. Equation (\ref{eq:ap}) defines this metric. $AP$ describes the precision-recall curve, where $P$ denotes precision and $R$ denotes recall, as a weighted mean. The weighted mean is calculated over a sum at each threshold $n$. We have chosen it as our evaluation metric because it also has been used in literature for SSL dataset evaluations\cite{wang2023ssl4eo}.

\begin{equation}\label{eq:ap}
	AP=\sum_{n}(R_n-R_{n-1})P_n
\end{equation}

\section{Results}
\label{sec:res}

We report and analyze the outcomes of the 12 experimental evaluations described in Section \ref{sec:exp}. Quantitative results for all experiments are summarized in Table \ref{tab:res}. Below we highlight the principal observations and interpret their implications.

\begin{table*}[]
	\centering
	\caption{Experimental results. Where Finetuned on described if cloudy or cloudless (denoted as BigEarthNet) has be used for finetuning. Tests have been also performed on cloudy and cloudless data}
	\begin{tabular}{@{}llll@{}}
		\toprule
		SSL Method         & Finetuned on  & Tested on BigEarthNet & Tested on CloudyBigEarthNet (Ours) \\ \midrule
		Moco2              & BigEarthNet & 79.620                  & 51.506           \\
		Moco2              & CloudyBigEarthNet (Ours) & 75.285                  & 72.502           \\
		Moco3              & BigEarthNet & 60.872                  & 37.067           \\
		Moco3              & CloudyBigEarthNet (Ours) & 54.834                  & 54.297           \\
		Masked Autoencoder & BigEarthNet & 73.726                  & 40.366           \\
		Masked Autoencoder & CloudyBigEarthNet (Ours) & 69.747                  & 69.087          
	\end{tabular}
	\label{tab:res}
\end{table*}

\paragraph{Performance In Cloud Free Test Cases}\label{para:cloudfree}
Models fine-tuned on our cloudy BigEarthNet variant remain competitive when evaluated on the original (cloud free) test set, but they typically show a modest reduction in performance relative to models fine-tuned on cloud free data. The largest single drop on the cloud free test set is 6.04 percentage points. This small penalty indicates that models adapted to cloudy data do not simply overfit to cloud artifacts, rather, they retain the ability to operate on cloud free images while gaining robustness to occlusions. 

The drop in accuracy is however, expected and caused by an implicitly learned modality preference. Prior works have shown in extensive ablation studies that optical information provides the most discriminative features for neural networks. When both modalities are combined, the multi-modal model yields the best overall performance, while in uni-modal cases, optical only models generally outperform radar only models \cite{wang2023ssl4eo,shen2023evaluating,he2025optical,crowson2019comparison}. Clouds, however occlude portions of the optical signal and reduce its utility. Consequently, models trained on cloudy data implicitly increase reliance on the cloud insensitive radar modality. This reweighting toward radar, while beneficial under cloudy conditions, explains the modest decrease in accuracy on cloud free evaluations, since radar is less informative than the full optical signal. The difference in extractable information is due to fewer amounts of available bands where our radar data only provides 2 bands while optical data provides 13. These results also support the validity of our cloudy dataset labels because our competent classifier has been proven to reliably detect the dominant tile-level classes.

\paragraph{Performance In Cloudy Test Cases}
When a model is fine-tuned exclusively on cloud free data and then evaluated on cloudy images, performance degrades substantially. The smallest observed drop in AP in this setting is 23 percentage points (best case). Across methods the relative performance gap between cloud free trained models and cloudy-finetuned counterparts ranges from 17.23 to 28.72 percentage points. These declines render cloud free trained models inferior to their cloudy-finetuned counterparts on realistic occluded imagery and underscore the importance of including cloudy samples during adaptation. 

A key finding is that models fine-tuned on the cloudy dataset perform similarly whether evaluated on cloud free or cloudy test sets. For MoCo v3 and MAE the difference in performance between these test sets is under 1 percentage point and for MoCo v2 the drop is 2.78 percentage points. This stability demonstrates that our finetuning procedure yields models that are robust to realistic cloud conditions, an important property for time critical applications (e.g. disaster response) where waiting for cloud free imagery is infeasible. For example flooding catastrophes are often accompanied by clouds and require therefore methods that perform well independent of cloud appearance. 

Even though, when finetuning on cloudy or cloudless data, both datasets had access to the same radar modality, only the finetuning on cloudy data was able to utilize it in cloudy imagery. Therefore, unless explicitly trained for it, a model doesn't learn to overcome the cloud problem even though it has the technical means for it by using the radar modality. We also want to highlight, that our realistic cloud dataset is only realistic and not a pure cloud challenge dataset as shown in Section \ref{sec:data}. This means, that while some parts of the data are cloudy, other parts are still cloud free, which let the cloud free trained approaches perform well on. 

\paragraph{Performance Analysis of SSL Approaches}
Masked Autoencoder (MAE) did not yield robustness to cloudy images despite its reconstructive, masking-based SSL objective, even though this masking based objective is similar to the occluding behavior of clouds. We hypothesize two causes. 

First, MAE typically applies uniform, random patch masking during pretraining and such a sampling strategy exposes evenly distributed visible patches that facilitate reconstruction of the entire scene, as it can be expected that one patch is available per region. In contrast, clouds form contiguous occlusions that mask whole connected regions. This behavior corresponds to a distribution that is not covered by uniform random patch masks. 

Second, He et al. has proposed and evaluated MAE on natural images which contain a single dominant concept per image, whereas in contrast remote-sensing tiles often contain multiple co-occuring land-cover classes\cite{he2022masked}. Therefore, a cloud that completely covers an instance can fully remove class defining features in a way that MAE pretraining does not model. These observations indicate that current masking strategies are not able to make reconstructive SSL robust to clouds, even though they behave similar, further highlighting the necessity for cloudy data in training sets. 

We observe a higher performance of MoCo v2 compared to MoCo v3 in our setup. One plausible explanation is that MoCo v3 abandons the explicit negative key queue and instead relies on very large batch sizes to provide diverse negatives \cite{chen2021empirical}. Our experiments use a batch size of 256, which is small relative to the regime where MoCo v3 was shown to excel. Accordingly, the momentum encoder and queue mechanism retained in MoCo v2 may be advantageous in our configuration.  

\paragraph{Practical Implications And Limitations}

Our results show that a backbone pretrained on cloud free data can be successfully adapted to cloudy downstream tasks by finetuning on a relatively small cloudy dataset. This suggests a pragmatic pathway for reusing existing high-quality SSL backbones in operational settings where cloud robustness is required. However, our downstream task uses tile-level, \textit{static} labels, as explained in Section \ref{sec:data} (e.g. arable land or urban fabric), that are tolerant to modest temporal shifts. The same adaptation strategy is not applicable for dynamic tasks such as object detection where labels change rapidly and which require pixel level precision.

\paragraph{Future Work}

Based on our findings, future directions should focus on attention mechanisms, including cloudy data in SSL datasets and defining cloud specific SSL pretext tasks. Attention that can localize cloud cover and selectively put a high weight value to cloud free pixels will improve the performance of methods finetuned on cloudy data when they are applied on cloud free data, as it was observed in Section \ref{sec:res}. Two ways to improve the SSL paradigm are identified. First, inclusion of cloudy samples directly in the SSL pretraining stage to learn representations that natively fuse radar and optical information under occlusions. Such advances could produce general-purpose SSL models that are robust to clouds without requiring task specific fine-tuning with cloudy data. Therefore, enhancing our current task specific approach to be generally applicable and enabling applications to tasks with non-static labels, which includes pixel-level precision tasks. Secondly, SSL objectives and masking strategies that explicitly exploit cloudy behavior and will therefore be applicable in a cloudy setting. 

\section{Conclusion}
\label{sec:conclusion}
We introduced the notion of \textit{static labels} as a practical criterion for creating a realistic, cloud distorted evaluation dataset in remote sensing. Leveraging this idea, we constructed and release a cloudy variant of BigEarthNet \cite{sumbul2021bigearthnet} that preserves tile-level (static) land-cover annotations while exposing models to realistic cloud occlusions (dataset and code: \url{https://github.com/mstricker13/CBEN}).

Following this contribution we have defined experiments to highlight the importance of including cloudy data in the training process. In summary, backbone networks have been trained in a self-supervised manner on the cloud free dataset SSL4EO-S12 \cite{wang2023ssl4eo}. Then, these backbones have been finetuned on two land use land cover classification datasets. First on the cloud free dataset BigEarthNet and a secondly on our cloudy version of BigEarthNet. 

Our experiments demonstrate two main findings. First, models fine-tuned only on cloud free data suffer substantial performance degradation when applied to cloudy imagery by at least 23 percentage points. Second, adapting models by fine-tuning on realistically cloudy examples yields robustness. Cloudy fine-tuned models maintain competitive performance on both cloud free and cloudy test sets and outperform cloud free trained counterparts on cloudy imagery by large margins between 17.23 and 28.72 percentage points.

These results have practical implications for operational earth-observation workflows. Including cloudy samples in fine-tuning processes is essential if models are expected to operate reliably under real world weather variability.

\bibliographystyle{IEEEtran}
\bibliography{ref}

\begin{IEEEbiography}[{\includegraphics[width=1in,height=1.25in,clip,keepaspectratio]{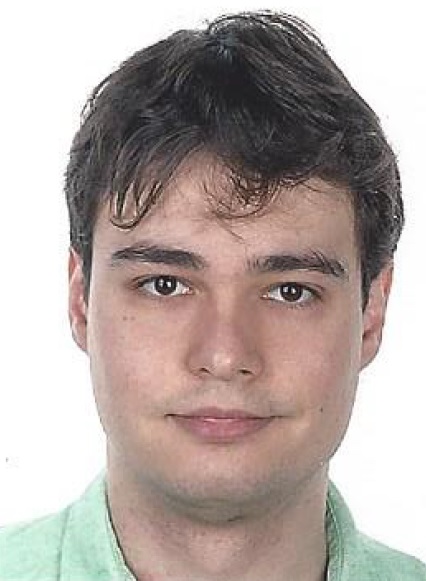}}]{Marco Stricker}
received the Bachelor of Science and Master of Science in computer science from Technical University Kaiserslautern, Kaiserslautern, Germany in 2017 and 2021. He is currently a Ph.D. candidate at the Osaka Metropolitan University, Osaka, Japan. His research interest are about applying machine learning in various applications, such as remote sensing. 
\end{IEEEbiography}

\begin{IEEEbiography}[{\includegraphics[width=1in,height=1.25in,clip,keepaspectratio]{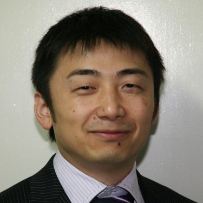}}]{Masakazu Iwamura}
	is a Professor at the Department of Computer Science and Intelligent Systems, Graduate School of Engineering, Osaka Metropolitan University. He received his B.E., M.E., and Ph.D. degrees in Engineering from Tohoku University, Japan, in 1998, 2000, and 2003, respectively. His research interests include text recognition, object recognition, and assistive technology for the visually impaired. He has received several awards, including IAPR/ICDAR Young Investigator Award in 2011, Best Paper Awards from IEICE in 2008 and 2021, Best Paper Awards from IAPR/ICDAR in 2007, IAPR Nakano Award (Best Paper Award) in 2010, ICFHR Best Paper Award in 2010, MVA Best Paper Award in 2017, and SIGACCESS Best Paper Award 2023. He served as the vice-chair of the IAPR Technical Committee 11 (Reading Systems) from 2016 to 2018, an Associate Editor of the International Journal of Document Analysis and Recognition from 2013 to 2023, and Associate Editor (2017–2021) and Associate Editor-in-Chief (2021–2023) of the IEICE Transactions on Information and Systems.
\end{IEEEbiography}

\begin{IEEEbiography}[{\includegraphics[width=1in,height=1.25in,clip,keepaspectratio]{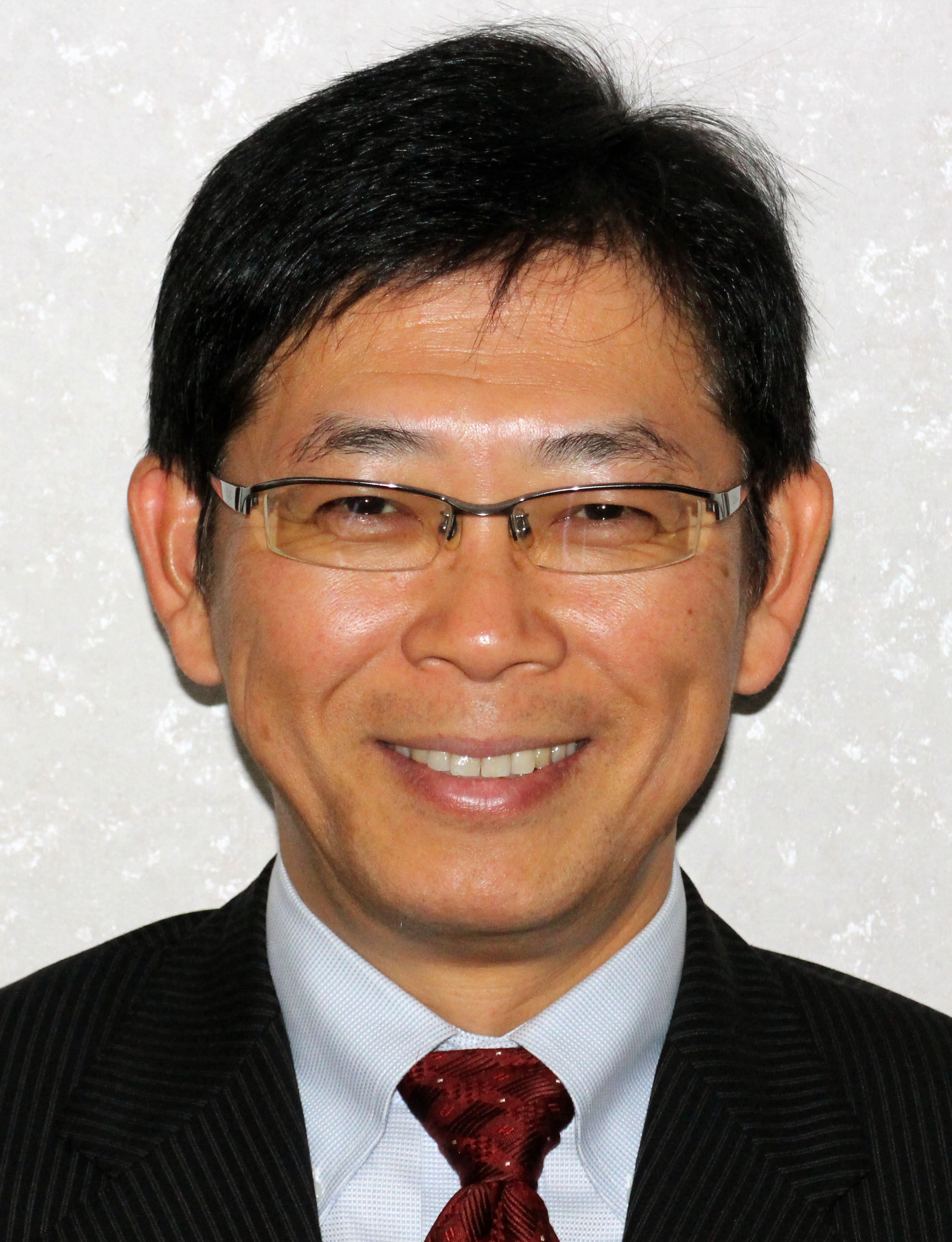}}]{Koichi Kise}
	received his Bachelor’s, Master’s, and Doctoral degrees in Communication Engineering from Osaka University, Japan, in 1986, 1988, and 1991, respectively. From 2000 to 2001, he was a visiting researcher at the German Research Center for Artificial Intelligence (DFKI), Germany. He is currently a Professor in the Department of Core Informatics at the Graduate School of Informatics, Osaka Metropolitan University, Japan.
	In 2008, he founded the Institute of Document Analysis and Knowledge Science (IDAKS) at Osaka Prefecture University (now Osaka Metropolitan University), where he continues to serve as Director. He is also the Director of the DFKI Lab Japan, established in 2022 at Osaka Metropolitan University as DFKI’s first overseas laboratory.
	He served as Vice Chair of IAPR TC11 from 2009 to 2012, and as Chair from 2012 to 2016. Since 2003, he has been a member of the Editorial Board of the International Journal of Document Analysis and Recognition (IJDAR), and he has served as its Editor-in-Chief since 2013. He gave invited talks at ICDAR 2015 and 2025, and was the General Chair of ICDAR 2017 held in Kyoto.
	He has received awards including best paper awards of three major international conferences in the field of document analysis, i.e., ICDAR (international conf. on document analysis and recognition), DAS (document analysis systems) and ICFHR (international conf. on frontiers in handwriting recognition). In 2025, he received the IAPR/ICDAR outstanding achievement award at ICDAR 2025.
	His research interests include document analysis, image understanding, human behavior understanding, learning augmentation, and AI applications in medicine.
\end{IEEEbiography}

\end{document}